\documentclass[runningheads]{llncs}
\usepackage{tikz}
\usepackage{longtable}
\usepackage{colortbl} % For table cell coloring
\usepackage{multirow}
\usepackage{siunitx} 
\usepackage[T1]{fontenc}
\usepackage{subcaption} 
\usepackage{caption}
\usepackage{graphicx}
\usepackage[ruled,vlined,linesnumbered]{algorithm2e}
\usepackage{hyperref}
\usepackage{xcolor}
\usepackage{siunitx} % for alignment of numbers

\usepackage{float}
\usepackage{amsmath}
\usepackage{cleveref}
\usepackage{booktabs}

\usepackage{color}
\begin{document}
%
% \title{Evaluating Large Language Models for Continual Relation Extraction in Incremental Task Learning}
\title{Post-Training Language Models for Continual Relation Extraction}
\titlerunning{Continual Relation Extraction}
\author{Sefika Efeoglu\inst{1}\orcidID{000-0002-9232-4840} \and
Adrian Paschke\inst{1,3}\orcidID{0000-0003-3156-9040} \and 
Sonja Schimmler\inst{2,3}\orcidID{0000-0002-8786-7250}}
\authorrunning{S. Efeoglu et al.}

\institute{
Freie Universität Berlin \and 
Technische Universität Berlin \and 
Fraunhofer Institute for Open Communication Systems, Kaiserin-Augusta-Allee 31, 10589 Berlin, Germany \\
\email{{sefika.efeoglu,adrian.paschke}@fu-berlin.de} \\
\email{sonja.schimmler@tu-berlin.de}
}

\maketitle              % typeset the header of the contribution
\begin{abstract}

% Real-world data, e.g., news articles, social media posts, and chatbot conversations, is dynamic and non-stationary, posing challenges for real-time structured representation via knowledge graphs (KGs). Relation extraction, essential for building KGs, struggles with adapting to dynamic data when traditional models are trained on outdated datasets. Continual Relation Extraction (CRE) methods address this by incremental learning of new relations while retaining prior knowledge. This paper explores using pre-trained language models for CRE, in particular large language models (LLMs), emphasizing memory replay to mitigate catastrophic forgetting. We evaluate decoder-only models (e.g., Mistral-7B and Llama2-7B) and encoder-decoder models (e.g., Flan-T5 Base) on TACRED and FewRel datasets. Incremental task fine-tuning of LLMs surpasses previous state-of-the-art approaches on TACRED in seen task accuracy and overall performance (in terms of whole and average accuracy), particularly with the Mistral and Flan-T5 models. Likewise, FewRel results outperform previous works in terms of seen task accuracies across incremental task learning. This work highlights critical aspects of knowledge transfer including positive backward knowledge transfer, advancing CRE with LLMs and memory replay for CRE in incremental task data.
Real-world data, such as news articles, social media posts, and chatbot conversations, is inherently dynamic and non-stationary, presenting significant challenges for constructing real-time structured representations through knowledge graphs (KGs). Relation Extraction (RE), a fundamental component of KG creation, often struggles to adapt to evolving data when traditional models rely on static, outdated datasets. Continual Relation Extraction (CRE) methods tackle this issue by incrementally learning new relations while preserving previously acquired knowledge. This study investigates the application of pre-trained language models (PLMs), specifically large language models (LLMs), to CRE, with a focus on leveraging memory replay to address catastrophic forgetting. We evaluate decoder-only models (eg, Mistral-7B and Llama2-7B) and encoder-decoder models (eg, Flan-T5 Base) on the TACRED and FewRel datasets. Task-incremental fine-tuning of LLMs demonstrates superior performance over earlier approaches using encoder-only models like BERT on TACRED, excelling in seen-task accuracy and overall performance (measured by whole and average accuracy), particularly with the Mistral and Flan-T5 models. Results on FewRel are similarly promising, achieving second place in whole and average accuracy metrics. This work underscores critical factors in knowledge transfer, language model architecture, and KG completeness, advancing CRE with LLMs and memory replay for dynamic, real-time relation extraction.

\end{abstract}
\section{Introduction}
\label{sec:into}

Real-world data sources, e.g., news articles on pandemics, social media posts related to hate speech, and chatbot conversations, generate large volumes of data, necessitating real-time analytical approaches. To enable real-time data analysis, it is crucial to represent this data in a structured format, e.g., knowledge graphs (KGs)~\cite{Amit_2019}, which can handle dynamic content effectively. This representation process relies on Information Extraction, with core tasks including Entity Recognition and Relation Extraction (RE)~\cite{Grishman_2015}. However, non-stationary data—generated continuously from real-world sources—introduces unique challenges. Traditional approaches, developed using stationary data, might fail to uncover significant new knowledge due to their reliance on outdated datasets during the development phase.

\begin{figure}[H]
\caption{Illustration of incremental training on relation extraction  tasks, followed by model evaluation on relation extraction test tasks for seen (or historical) relation types.}
\label{fig:incremental_RE}
\centering\includegraphics[width=1.0\linewidth]{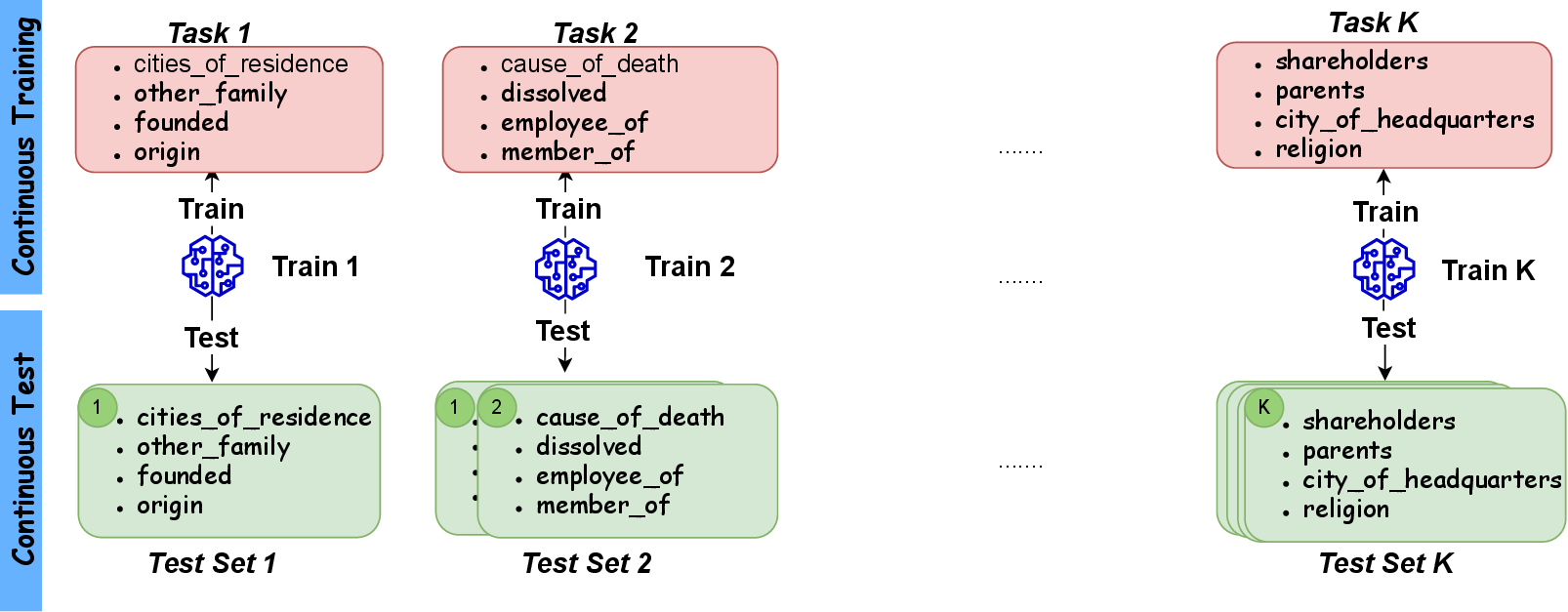}
\end{figure}

To process unstructured non-stationary data and represent it in KGs, continuous learning techniques can identify entities and their relationships while handling real-time streaming data. In real-world scenarios, an RE model trained on stationary data might fail to recognize new relation types not introduced during its training phase. Streaming data, which is inherently non-stationary, requires continuous training and evaluation to detect these new relation types.~\Cref{fig:incremental_RE} illustrates how a model is incrementally trained on new RE tasks and evaluated on an incrementally expanding test set of relation types.

Continual —also known as lifelong or incremental—learning approaches were initially introduced for computer vision problems and later extended to natural language processing and information extraction tasks such as RE~\cite{biesialska-etal-2020-continual}. Continual relation extraction (CRE) approaches are typically categorized into (i) architecture-based methods~\cite{CHEN2023110288,duan_2023}, (ii) memory replay-based methods~\cite{xia-etal-2023-enhancing,zhao-etal-2022-consistent}, and (iii) regularization-based methods (e.g., integrating feature regularization~\cite{CHEN2023110288,Van_Nguyen_2024,Hangjie_2020})  \cite{cui-etal-2021-refining}. These approaches aim to incrementally transfer knowledge from previous tasks to subsequent ones by applying incremental task learning techniques~\cite{van2018three}. Incremental fine-tuning of models on new tasks is a common approach in continual learning (CL). However, it faces the challenge of catastrophic forgetting, where the model loses knowledge of previously learned relation types. Although some attempts have been made to address catastrophic forgetting using architecture-based and regularization-based CRE methods~\cite{CHEN2023110288,Van_Nguyen_2024}, memory replay has emerged as the most promising approach. Memory replay-based methods, inspired by human-like learning in neuroscience~\cite{vandeven2024continuallearningcatastrophicforgetting}, utilize encoders, particularly BERT~\cite{chen-etal-2023-consistent,cui-etal-2021-refining,wang-etal-2019-sentence,ye-etal-2024-distilling,zhang_2022}, along with custom models with randomly initialized weights to select memory samples from previously learned tasks in previous works.

Recent advancements in large language models (LLMs) have brought decoder-only, encoder-only and encoder-decoder models to the forefront, achieving state-of-the-art performance in mainstream tasks such as entity recognition~\cite{shlyk2024real}, question answering~\cite{efeoglu2024large}, and traditional RE~\cite{efeoglu2024retrievalaugmented}. Additionally, Zhou et al.~\cite{ijcai_cl_2024} recommend that pre-trained model-based CL might outperform traditional CL approaches that rely on randomly initialized weights, although their discussion focuses primarily on computer vision applications.
In this paper, we aim to address the following open research question: \textit{To what extent do pre-trained language models affect knowledge transfer, including backward transfer, in continual relation extraction?}

To the best of our knowledge, no previous work has evaluated LLMs for CRE while mitigating the catastrophic forgetting problem in incrementally fine-tuned LLMs using memory replay. To achieve this, we utilize Flan-T5 Base~\footnote{Flan-T5 Base~\url{https://huggingface.co/google/flan-t5-base}, accessed date: 10.03.2024}, Mistral-7B-Instruct-v2.0~\footnote{Mistral-7B-Instruct-v2.0:\url{https://huggingface.co/mistralai/Mistral-7B-Instruct-v0.2}, accessed date: 10.03.2024}, and Llama2-7b-chat-hf\footnote{Llama2-7b-chat-hf:\url{https://huggingface.co/meta-llama/Llama-2-7b-chat-hf}, accessed date: 10.03.2024} along with memory replay. We apply K-means clustering for selecting memory samples on the TACRED~\cite{zhang-etal-2017-position} and FewRel~\cite{han-etal-2018-fewrel} benchmarks. The outcomes are as follows:
\begin{itemize}  
    \item \textbf{Outstanding Performance in Incremental Task Learning for CRE}: Both Flan-T5 and Mistral outperform previous state-of-the-art method on TACRED and FewRel, achieving higher seen task accuracy at the end of incremental task learning process. Furthermore, Mistral achieves the highest performance among the three language models used in this work.

    \item \textbf{Backward Knowledge Transfer}:  
    \begin{itemize}  
        \item Llama2 exhibits positive backward knowledge transfer on both TACRED and FewRel. This positive knowledge transfer helps reduce hallucinated predictions in earlier-learned tasks.
        \item Mistral demonstrates positive backward knowledge transfer on TACRED, while  observing a slight forgetting issue on FewRel.
    \end{itemize}  

    \item \textbf{Challenges with FewRel}: Flan-T5 Base experiences significant catastrophic forgetting on FewRel, likely due to its shorter average token count per sentence (25.0 on FewRel compared to 34.2 on TACRED).
\end{itemize}

\noindent In the rest of this paper, we first provide an overview about the preliminaries and related works in~\Cref{sec:related_works}. We then introduce our methodology for CL of new relation types in~\Cref{sec:method}. Following this, we evaluate our approach using two benchmark datasets and various LLMs, applying memory replay with incremental task learning, and compare it with previous methods in~\Cref{sec:evaluation}, and then conduct ablation study in~\Cref{sec:ablation_study}. Following this, we discuss potential outcomes of our work with previous approach within the context of our research question in~\Cref{sec:discussion}. Finally, we conclude by highlighting our findings and suggesting a future research direction in~\Cref{sec:conclusion}.

\section{Related Works} \label{sec:related_works}
Continual relation extraction (CRE) is a task that aims to train a model continuously on data containing new relation types, while preserving the knowledge of previously learned relations~\cite{xia-etal-2023-enhancing,zhao-etal-2022-consistent} as illustrated in~\Cref{fig:incremental_RE}. CRE task might be formalized by definitions based in~\cite{Zhang_fprompt_2024}. Furthermore, \noindent an example of data for a relation type is defined as a tuple, $\mathbf{x = \langle \textit{sentence}, \textit{head}, \textit{tail} \rangle}$  where \textit{sentence} is a textual sentence consisting of multiple tokens, \textit{head} is the token(s) corresponding to the head entity, and \textit{tail} is the token(s) corresponding to the tail entity. \noindent Additionally, $\textit{Data sequence}= \{X_0, X_1, \dots, X_K\}$ and $\textit{Relation sequence}= \{R_0, R_1, \dots, R_K\}$
where $X_K$ contains the tuples for relations $R_t$ at time $t \leq K$, with $t$ indicating the time step. CRE approaches are commonly categorized into three types: (i) memory replay-based methods, (ii) dynamic architecture-based methods and (iii) regularization-based methods. In this section, we detail previously proposed methods designed to address the catastrophic forgetting phenomenon in CRE. The most commonly used benchmark datasets for CRE are TACRED~\cite{zhang-etal-2017-position} and FewRel~\cite{han-etal-2018-fewrel}.

 \textbf{Memory Replay-Based Methods.} 
This approach utilizes a memory buffer to store a limited number of samples, which are replayed after training on each new task in the context of continual learning (CL). Wang et al.~\cite{wang-etal-2019-sentence} propose a sentence alignment model integrated with simple memory replay for task-incremental relation extraction, a technique referred to as CRE. Building on this, Cui et al.~\cite{cui-etal-2021-refining} introduce a prototypical framework to refine sample embeddings stored in memory for replay, alongside relation prototypes. Furthermore, Chen et al.~\cite{chen-etal-2023-consistent} tackle the catastrophic forgetting problem by employing a consistency learning module designed to mitigate distributional shifts between old and new tasks in a few-shot CL. More recently, Zhang et al.~\cite{zhang_2022} propose the Knowledge-Infused Prototypes (KIP) framework, which leverages multi-head scaled dot-product attention to integrate features derived from relational knowledge-infused prompts, distinguishing it from other prototype-based methods. In contrast, Ye et al.~\cite{ye-etal-2024-distilling} address the dual challenges of limited labeled data and data imbalance. Their approach employs causal inference to effectively select and store memory samples for few-shot CRE.

\textbf{Architecture-Based Methods.}
Duan et al.~\cite{duan_2023} propose a zero-shot relation representation method that uses instance prompting and prototype rectification to refine relation instance and prototype embeddings simultaneously. Additionally, Chen et al.~\cite{CHEN2023110288} introduce a three-phase learning strategy—preliminary learning, memory retention, and memory reconsolidation—enhanced by linear connectivity to balance plasticity and stability.

\textbf{Regularization-Based Methods.}
Hangjie et al.~\cite{Hangjie_2020} propose a dynamic feature regularization approach that calculates dynamic loss during the training process to mitigate the catastrophic forgetting problem. Similarly, Jialan et al.~\cite{jialan_2023} employ an LSTM architecture with backward projection to preserve the classification space for relation types. In another work, Wu et al.~\cite{wu_fei_2024} integrate contrastive learning with a prompt-based BERT encoder, advancing few-shot CRE. As opposed to encoder-based methods, Le Nguyen et al.~\cite{Van_Nguyen_2024} propose a gradient-based sequential multi-task approach for CRE that addresses multi-objective training in CL without requiring encoder retraining.

Leveraging advancements in LLMs, recent efforts have explored their use for CRE. Tirsogoiu et al.~\cite{tirsogoiu2023learned} evaluate generative models for relation type identification, comparing clustering performance across zero-shot, one-shot, and few-shot settings. Xiong et al.~\cite{xiong-etal-2023-rationale} propose contrastive rational learning with prompting to improve CL. In this work,  we evaluates pre-trained LLMs for task-incremental relation extraction using memory replay and instruction fine-tuning, assessing Flan-T5 Base, Llama2-chat-7b-hf, and Mistral-Instruct-v2.0 on TACRED and FewRel datasets.
% \newpage
\section{Methodology}
\label{sec:method}
In this section, we present our methodology, integrating incremental task fine-tuning of pre-trained language models (PLMs) with memory replay techniques (see~\Cref{sec:t-inc-finetune}). The PLMs are fine-tuned using prompt instruction datasets derived from the original datasets through a structured prompt template selection process detailed in~\Cref{sec:prompt_template}.

\subsection{Continuous Learning with PLMs}
\label{sec:t-inc-finetune}
In this work, we continuously train PLMs on a stream of incoming tasks, \(T_1, T_2, \dots,T_K\), followed by the application of memory replay to the model after training on a new task. Memory replay is applied to the model after each subsequent task.

The continuous fine-tuning process follows the steps outlined in~\Cref{algo:task_inc_train}. As discussed in~\Cref{sec:into}, memory replay (Lines 9–10 in~\Cref{algo:task_inc_train}) is among the most effective strategies for mitigating catastrophic forgetting in continual learning. To select samples from the training data of previous tasks for memory replay after training on a new task (Lines 3–6), the most representative samples are identified by applying K-means clustering to the centroids of the clusters. This process utilizes embeddings from either the encoder or decoder of the trained model, depending on the architecture of the PLM. Specifically, decoder embeddings are used for decoder-only models, whereas encoder embeddings are applied for encoder-decoder models. After completing training on the new task, the validation dataset is used to optimize training parameters—such as the learning rate—based on validation loss (Lines 5–6) and this step is repeated throughout training (Lines 3-6). Validation datasets were not utilized for optimization in previous works~\cite{cui-etal-2021-refining,zhao-etal-2022-consistent}.

\begin{algorithm}[H]

\SetAlgoNlRelativeSize{-1} % Reduce line numbering font size
\SetAlgoLined
\caption{Incremental Task Instruction Fine-Tuning for Pre-trained Language Models.}
\label{algo:task_inc_train}
\KwIn{Stream of tasks $T_{1}, T_{2},\dots$, memory samples $\widetilde{M} \gets \emptyset$, pre-trained language model $f_{\theta}$ where $\theta$ is the model weights, memory size $m$}
\KwOut{Relation Classification Model $\widetilde{f_{\theta}}$}

\While{there are still tasks}{
    Retrieve current task $T_{k}$; 
    /* $k$: position of current task */

    \For{$i \gets 1$ \KwTo $epoch1$}{
        Update $\theta$ with $\nabla L$ on ${D}^{k}_{train}$; 
        /* $\nabla L$: gradient of classification loss on ${D}^{k}_{train}$ for $T_k$ */

        Evaluate the model on $D_{valid}^k$ and compute the validation loss $L_{valid}$;

        Adjust the learning rate based on $L_{valid}$;
    }

    Select $m$ memory samples $M_k$ from $D_{train}^k$ using K-means per relation type;
    /* Use K-means to select representative samples */

    \For{$i \gets 1$ \KwTo $epoch2$}{
        Update $\theta$ with $\nabla L$ on $\widetilde{M}$; 
        /* Fine-tune on memory $\widetilde{M}$ */
    }

    $\widetilde{M} \gets \widetilde{M} \cup M_{k}$; 
    /* Add selected samples from $T_{k}$ to memory */
}
\end{algorithm}
In reference to the details in~\Cref{algo:task_inc_train}, the algorithm processes a stream of tasks and memory samples alongside the PLM model ($f_{\theta}$). The memory samples ($\widetilde{M}$) are initially empty and are dynamically selected from the training dataset (${D}^{k}_{train}$) to facilitate replay after training on the new task. The PLM model is incrementally trained on ${D}^{k}_{train}$ and validated on ${D}^{k}_{valid}$ per task as it arrives from the stream (Lines 3-6). Following the training of the PLM on the new task, memory samples are selected from the training data using K-means clustering (see Line 8). This selection leverages either the PLM's encoder ($f_{\theta}^{encoder}$) or decoder to compute embeddings of the samples, previously trained in Lines 3–6. Before storing the selected samples, which represent the centroids of the clusters identified by K-means (Line 12), the memory samples from the previous task, denoted as $\widetilde{M}$, are replayed in Lines 9-10.

\subsection{Prompt Template Selection}
~\label{sec:prompt_template}
\begin{figure}[H]
    \centering
    % Subfigure 1
    \begin{subfigure}[t]{\textwidth}
        \centering
        \includegraphics[width=1.0\linewidth]{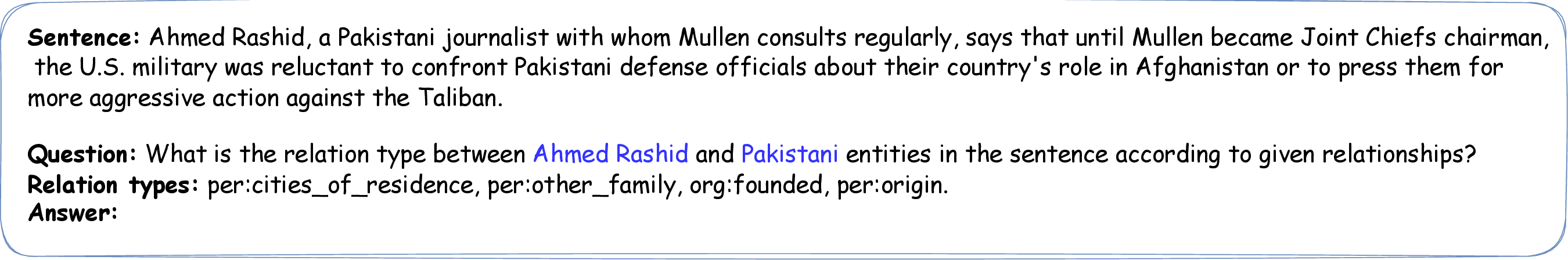}
        \caption{An Example for Prompt Template 1.}
        \label{fig:prompt_one}
    \end{subfigure}
    \vspace{1em} % Space between subfigures
    % Subfigure 2
    
    \begin{subfigure}[t]{\textwidth}
        \centering
        \includegraphics[width=1.0\linewidth]{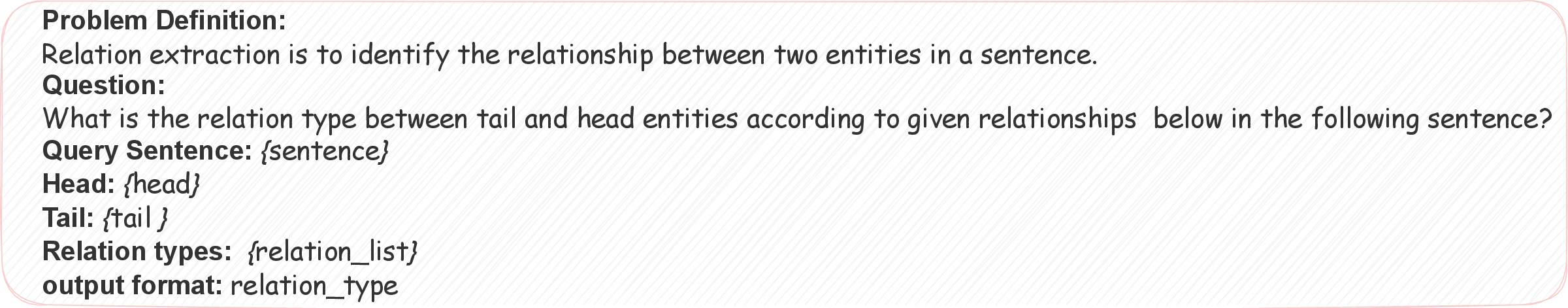}
        \caption{Prompt template 2.}
        \label{fig:prompt_two}
    \end{subfigure}
    \caption{(a) Modified version of the prompt template in~\cite{tran-etal-2024-preserving} with entities highlighted. (b) Prompt template 2 from~\cite{efeoglu2024retrievalaugmented}. The relation types dynamically change according to the task-specific relation types.}
    \label{fig:combined_prompts}
\end{figure}

We investigate two prompt templates for incremental task fine-tuning without memory replay. The first template (see~\Cref{fig:prompt_one}), derived from~\cite{tran-etal-2024-preserving}, is modified to incorporate relation types and employs conditional generation techniques~\cite{madaan-etal-2022-conditional}. In contrast, the second template are taken from~\cite{efeoglu2024retrievalaugmented}, explicitly defines the task and specifies the head and tail entities, differentiating it from Template 1 (see~\Cref{fig:prompt_one}). We then assess the model's performance with both prompt templates, presenting the results in~\Cref{sec:evaluation}. We consider the best-performing template for subsequent experiments based on these results.

\section{Evaluation}
\label{sec:evaluation}
In this section, we first outline the experimental settings in~\Cref{sec:experimental_settings} and then present the results in~\Cref{sec:results} and knowledge transfer analysis in~\Cref{sec:kta}, based on the performance metrics described in this section.
\subsection{Experimental Settings}
\label{sec:experimental_settings}
\noindent \textbf{a.) Datasets.}
We evaluate our approach using two benchmark datasets, TACRED~\cite{zhang-etal-2017-position} and FewRel~\cite{han-etal-2018-fewrel}, which are used for continual relation extraction. Due to the imbalance of TACRED, we follow the experimental settings outlined in prior work~\cite{cui-etal-2021-refining,zhao-etal-2022-consistent}, excluding the \textit{no\_relation} class to tackle this issue. For each remaining relation type, we randomly select up to 320 sentences for training and 40 sentences each for validation and testing in each task. We incrementally train our models on TACRED, which is divided into ten tasks, each containing four relation types. For FewRel, we use the same settings as in~\cite{cui-etal-2021-refining,zhao-etal-2022-consistent}, where each task includes eight relation types across ten tasks. For each relation type, we randomly select 420 sentences for training and 140 sentences for validation and test in each task. To ensure consistency, we use the same relation type combinations as those in the published results of ~\cite{zhao-etal-2022-consistent} from their open-source repository and conduct our experiments over five runs with a memory sample size of 10, which is considered ideal in~\cite{cui-etal-2021-refining,zhang_2022,zhao-etal-2022-consistent}.
% \begin{figure}[H]
%     \centering
%     \includegraphics[width=0.4\linewidth]{images/data_stats.pdf}
%     \caption{Statistics about number of tokens per sentence in both datasets.}
%     \label{fig:data_stats}
% \end{figure}

\noindent \textbf{b.) Pre-Trained Language Models.}
We employ three pre-trained language models with distinct architectures: Flan T5 Base, Mistral-7b-Instruct-v2.0, and Llama2-7b-chat-hf. As prior continual relation extraction approaches mainly used encoder-only models like BERT, we exclude encoder-only models in this work. The reason why is that we utilize fine-tuned versions of Llama2-7b and Mistral-7b in the work, as Flan T5 has also been trained on instruction-based tasks~\cite{qorib-etal-2024-decoder}. The default cross-entropy loss is used in all of these experiments.

\noindent \textbf{c.) Evaluation Metrics.} The performance of our experiments are evaluated according to seen task accuracies across incremental task learning (ITL) as illustrated in~\Cref{fig:incremental_RE}. Additionally, we also computed whole accuracy, average accuracy and backward knowledge transfer metrics. \textbf{Whole Accuracy}~\cite{zhang-etal-2017-position} is computed from the resulting model at the end of ITL on all test data of all tasks. \textbf{Average Accuracy}~\cite{zhang-etal-2017-position} is also computed from the resulting model trained on task k on all the test sets of all tasks seen up to stage k of ITL. \textbf{Backward Knowledge Transfer}~\cite{pmlr-v235-hu24b} quantifies the degree of forgetting in previously learned tasks after learning a new one. This metric is crucial for determining whether backward knowledge transfer (\textit{bwt}) occurs~\cite{vandeven2024continuallearningcatastrophicforgetting}. Notably, no prior work in continual relation extraction has reported on this metric~\cite{vandeven2024continuallearningcatastrophicforgetting}, which is computed by
$bwt = \frac{1}{N-1}\sum_{t=1}^{N} A_{N,t}-A_{t,t}$
 where $A_{N,t}$ represents the test accuracy on the $t$-th task after sequential training on all $N$ tasks. All metrics results in this work are the mean of five runs, except for results shown in confusion matrices.

\noindent \textbf{d.) Parameter Settings.} The parameters are for the best performing models trained on A100 40 GB GPU memory with Colab.
Throughout the model training process, LoRA~\cite{hu2021lora} is applied to 4-bit quantized pre-trained language models (QLoRA~\cite{dettmers2023qlora}) to minimize GPU requirements while focusing on the targeted modules of the language models.
\noindent \textbf{Model Parameters.} For Flan T5 Base on TACRED, we use epochs: 5, batch size: 8, learning rate: 0.001, and a cosine scheduler. For FewRel, the batch size and epochs are 16 and 5, respectively. For decoder-only models Mistral and Llama2, we use 5 epochs, batch size: 4, weight decay: 0.001, learning rate: 0.002, and a cosine scheduler on TACRED, with 5 epochs and a batch size of 8 on FewRel.
\noindent \textbf{LoRA Parameters.}
$LoRA_{\alpha}$ is set to 32, the rank parameter is 4, the task type is Seq2SeqLM, and $LoRA_{dropout}$ is 0.01 for the Flan T5 Base model. For Mistral and Llama2, $LoRA_{\alpha}$ is set to 16, the rank parameter is 64, the task type is CausalLM, and $LoRA_{dropout}$ is 0.1.

\noindent \textbf{e.) Prompt Template Selection.}
We compare two prompt templates for incremental task fine-tuning on TACRED across five runs. The results on the TACRED dataset indicate that for prompt type one, mean Whole Accuracy (\textit{w}) is 92.7\%, and the mean Average Accuracy (\textit{a}) is 92.1\%. In comparison, prompt type two achieves mean \textit{w} of 90.5\% and mean \textit{a} of 90.6\%. Therefore, we take into account prompt template one (see~\Cref{fig:prompt_one}) to create prompt datasets from TACRED and FewRel benchmarks to fine-tune the aforementioned pre-trained language models.

\subsection{Results}
\label{sec:results}
We evaluate different versions of three well-known large language models—Flan T5 Base, Llama2-7b-chat-hf, and Mistral-7b-Instruct-v2.0—alongside incremental task learning (ITL) utilizing memory replay, with a memory sample size of 10,  for continual relation extraction. We leverage two widely used benchmark datasets: (i) TACRED and (ii) FewRel, conducting experiments five times for each.
\sisetup{table-format=3.1} %
\begin{table}[H]
\tiny
\caption{The models trained on corresponding tasks are evaluated on test datasets of previously seen tasks across incremental task learning.}
\label{tab:results}
\centering
\begin{tabular}{@{}lcccccccccc@{}}

\toprule
\multicolumn{11}{c}{\textbf{TACRED}}\\
\cmidrule(lr){1-11}
\textbf{Method} & \multicolumn{10}{c}{\textbf{Index of Tasks for Base Training}} \\

\cmidrule(lr){2-11}
&  \textbf{1} & \textbf{2} & \textbf{3} & \textbf{4} & \textbf{5} & \textbf{6} & \textbf{7} & \textbf{8} & \textbf{9} & \textbf{10} \\ 
\midrule
% \multirow{7}{*}{\rotatebox{90}{\textbf{TACRED}}} 

 \textbf{(published SoTA) KIP-Framework}~\cite{zhang_2022} & \cellcolor{cyan!25}98.3 & 95.0 & 90.8 & 87.5 & 85.3 & 84.3 & 82.1 & 80.2 & 79.6 & 78.6 \\
\cmidrule{1-11}

 \textbf{Ours with Flan-T5 Base} & 96.0 & \cellcolor{cyan!25}96.2 & 95.7 & \cellcolor{cyan!25}96.0 & 95.7 & 95.4 & 96.0 & 96.0 & \cellcolor{cyan!25}96.3 & 95.8 \\
  \textbf{+ Mistral} & 95.0 & 94.8 & \cellcolor{cyan!25}96.4 & \cellcolor{cyan!25}96.0 & \cellcolor{cyan!25}96.6 & \cellcolor{cyan!25}97.0 & \cellcolor{cyan!25}96.8 & \cellcolor{cyan!25}96.9 & 95.8 & \cellcolor{cyan!25}96.9 \\
 \textbf{+ Llama2} &  57.7 & 57.6 & 54.9 & 55.8 & 57.6 & 62.0 & 62.4 & 65.3 & 67.7 & 70.6 \\
\\
\midrule
\multicolumn{11}{c}{\textbf{FewRel}}\\
\cmidrule(lr){1-11}
% \multirow{7}{*}{\rotatebox{90}{\textbf{FewRel}}} 

% & \textbf{EMAR}~\cite{han-etal-2020-continual} & 88.5 & 73.2 & 66.6 & 63.8 & 55.8 & 54.3 & 52.9 & 50.9 & 48.8 & 46.3\\
% &\textbf{EA-EMR}~\cite{wang-etal-2019-sentence} & 88.5 & 69.0 & 59.1 & 54.2 & 47.8 & 46.1 & 43.1 & 40.7 & 38.6 & 35.1\\

% & \textbf{EMAR-BERT} & \cellcolor{cyan!25}98.8 & 89.1 & 89.5 & 85.7 & 83.6 & 84.8 & 79.3 & 80.0 & 77.1 & 73.8\\
\textbf{(published SoTA) KIP-Framework}~\cite{zhang_2022} & \cellcolor{cyan!25}98.4 & 93.5 & 92.0 & 91.2 & 90.0 & 88.2 & 86.9 & 85.6 & 84.1 & 82.5\\
\cmidrule{1-11}
 \textbf{Ours with Flan-T5 Base} & 96.70 & \cellcolor{cyan!25}94.83 & \cellcolor{cyan!25}95.12 & 93.47 & 93.23 & \cellcolor{cyan!25}92.40 & 91.38 & 91.69 & \cellcolor{cyan!25}91.04 & 89.58 \\
\textbf{+ Mistral} & 95.98    & 94.61 & 94.71 &\cellcolor{cyan!25} 93.56 &\cellcolor{cyan!25} 93.57 & 92.26 &\cellcolor{cyan!25} 92.46 &\cellcolor{cyan!25} 91.91 &72.91 &\cellcolor{cyan!25} 91.35 \\
 \textbf{+ Llama2} &15.42 & 27.77 & 38.88 & 44.24 & 52.13 & 57.44 & 62.18 & 67.73 & 69.38 & 71.29  \\
\\
\bottomrule
\end{tabular}
\end{table}
The ITL-based fine-tuned Flan T5 Base model achieves remarkable performance on TACRED, achieving impressive mean seen task accuracy with 95.8\% on TACRED and with 89.58\% on FewRel as shown in~\Cref{tab:results}. It also demonstrates strong whole accuracy (\textit{w}) and average accuracy (\textit{a}), dealt with minimal forgetting on TACRED, as indicated by the mean backward knowledge transfer (\textit{bwt}) of -0.2\% in~\Cref{tab:w_a_acc}. Unfortunately, the model encounters significant forgetting challenges on FewRel, with a mean \textit{bwt} of -1.75\% (see~\Cref{tab:w_a_acc}). Furthermore, the mean \textit{w} and \textit{a} are 95.76\% and 95.78\% on TACRED, and 89.61\% and 89.61\% on FewRel, respectively as given in~\Cref{tab:w_a_acc}, indicating good performance on individual tasks with minimal catastrophic forgetting on TACRED. 

In addition to Flan T5 Base, we evaluate the performance of Mistral on TACRED and FewRel as well. Mistral achieves positive \textit{bwt} with 0.17\% on TACRED in~\Cref{tab:w_a_acc}, indicating that it enhances the performance of previously learned tasks at the end of ITL. However, it encounters slight forgetting with a mean \textit{bwt} of -1.0\% on FewRel (see~\Cref{tab:w_a_acc}). Besides, the resulting models' mean seen task accuracies on these benchmark datasets are 96.9\% and 91.35\%, respectively in~\Cref{tab:results} at the end of ITL where Mistral is used. The mean \textit{w} and \textit{a} are 96.89\% and 96.76\% on TACRED, and 94.93\% and 94.93\% on FewRel with this model in~\Cref{tab:results}. Likewise, Mistral performs well on individual tasks with minimal catastrophic forgetting on TACRED when \textit{w} and \textit{a} metrics are considered.

Finally, we evaluate Llama2 in the context of ITL settings. In contrast to Flan T5 Base and Mistral, it does not achieve remarkable results on either dataset, with a mean seen task accuracy of 70.6\% on TACRED and 71.29\% on FewRel in~\Cref{tab:results} at the end of ITL. Interestingly, it demonstrates positive \textit{bwt} on both datasets (see~\Cref{tab:w_a_acc}), a phenomenon rarely observed even in computer vision. Similar to its seen task accuracies, its mean \textit{w} and \textit{a}—71.17\% and 70.86\% on TACRED, and 71.29\% for both metrics on FewRel—are lower than those of the other models (see~\Cref{tab:w_a_acc}), primarily due to hallucinating relation types by Llama2 like \textit{per:affiliate} and \textit{per:columnist}.

Consequently, Mistral achieves the best results on both datasets among the three language models, even though it encounters slight forgetting on FewRel. Furthermore, Flan T5 Base struggles with significant catastrophic forgetting depending on the dataset. Although Llama2 does not achieve performance comparable to Mistral and Flan T5 Base on either dataset, it demonstrates positive knowledge transfer on both datasets. The Llama2 and Flan T5 Base performance will be explored and discussed in the next section.
\begin{table}[H]
\centering
\tiny
\caption{Mean Average Accuracy (a), Whole Accuracy (w) (\%), and Backward Knowledge Transfer (\textit{bwt}) (\%) on TACRED and FewRel datasets over 5 runs. Second-best results are in green, and best results are in blue. `-' indicates no result for the metric.}
\label{tab:w_a_acc}
\sisetup{table-format=3.1} % Configures decimal alignment

\begin{tabular}{lccc|ccc|ccc}
\toprule
\textbf{Method} & \multicolumn{3}{c}{\textbf{TACRED}} & \multicolumn{3}{c}{\textbf{FewRel}} & \multicolumn{3}{c}{\textbf{Average}} \\
\cmidrule(lr){2-4} \cmidrule(lr){5-7} \cmidrule(lr){8-10} 
 & \textbf{w} & \textbf{a} & \textbf{bwt} & \textbf{w} & \textbf{a} & \textbf{bwt} & \textbf{w} & \textbf{a} & \textbf{bwt} \\
\midrule

\textbf{(published SoTA) KIP-Framework}~\cite{zhang_2022} & 91.1 & 91.6 & -- & \cellcolor{cyan!25}96.3 & \cellcolor{cyan!25}96.6 & -- & \cellcolor{green!25}93.7 & \cellcolor{green!25}94.1 & -- \\
\cmidrule(lr){1-10}
\textbf{Ours with Flan-T5 Base} & \cellcolor{green!25}95.76 & \cellcolor{green!25}95.78 & -0.20 & 89.61 & 89.61 &-1.75 & 92.68 & 92.7& -0.98 \\
\quad \quad \quad \textbf{+ Mistral-7b} & \cellcolor{cyan!25}96.89 & \cellcolor{cyan!25}96.76 & 0.17 & \cellcolor{green!25}94.93 & \cellcolor{green!25}94.93 &  -1.00 & \cellcolor{cyan!25}95.91 & \cellcolor{cyan!25}95.85 & -0.42 \\
\quad \quad \quad \textbf{+ Llama2-7b} &71.17 & 70.86 & 1.69& 71.29 & 71.29 & 6.08  & 71.5 & 71.08 & 3.89 \\
\bottomrule
\end{tabular}

\end{table}

\subsection{Knowledge Transfer Analysis}
\label{sec:kta}

\begin{figure}[H]
    \centering
    % Subfigure 1
    \begin{subfigure}[b]{0.4\textwidth}
        \centering
        \includegraphics[width=0.85\linewidth]{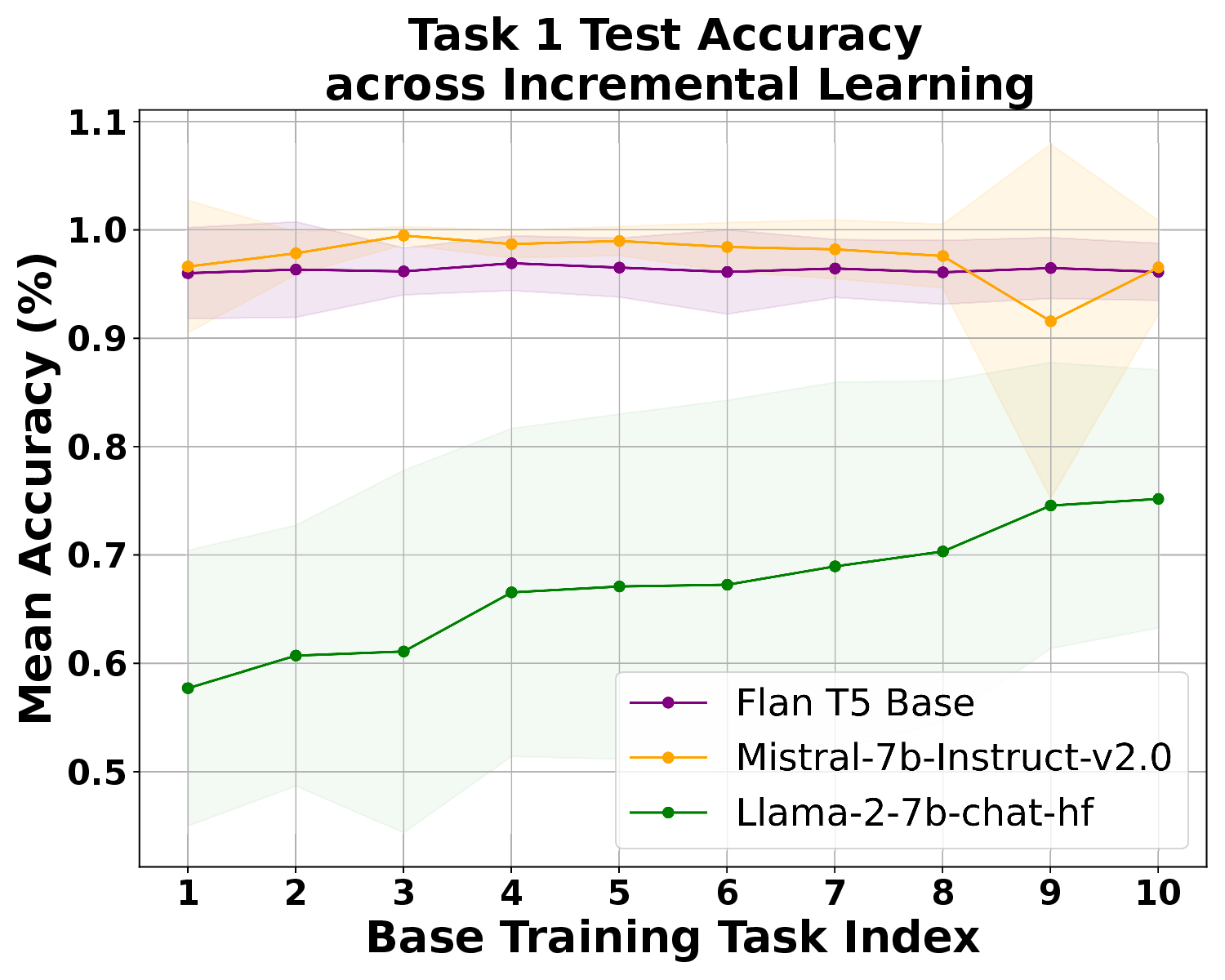}
        \caption{TACRED}
        \label{fig:tacred}
    \end{subfigure}
    % \hfill
    % Subfigure 2
    \begin{subfigure}[b]{0.4\textwidth}
        \centering
        \includegraphics[width=0.85\linewidth]{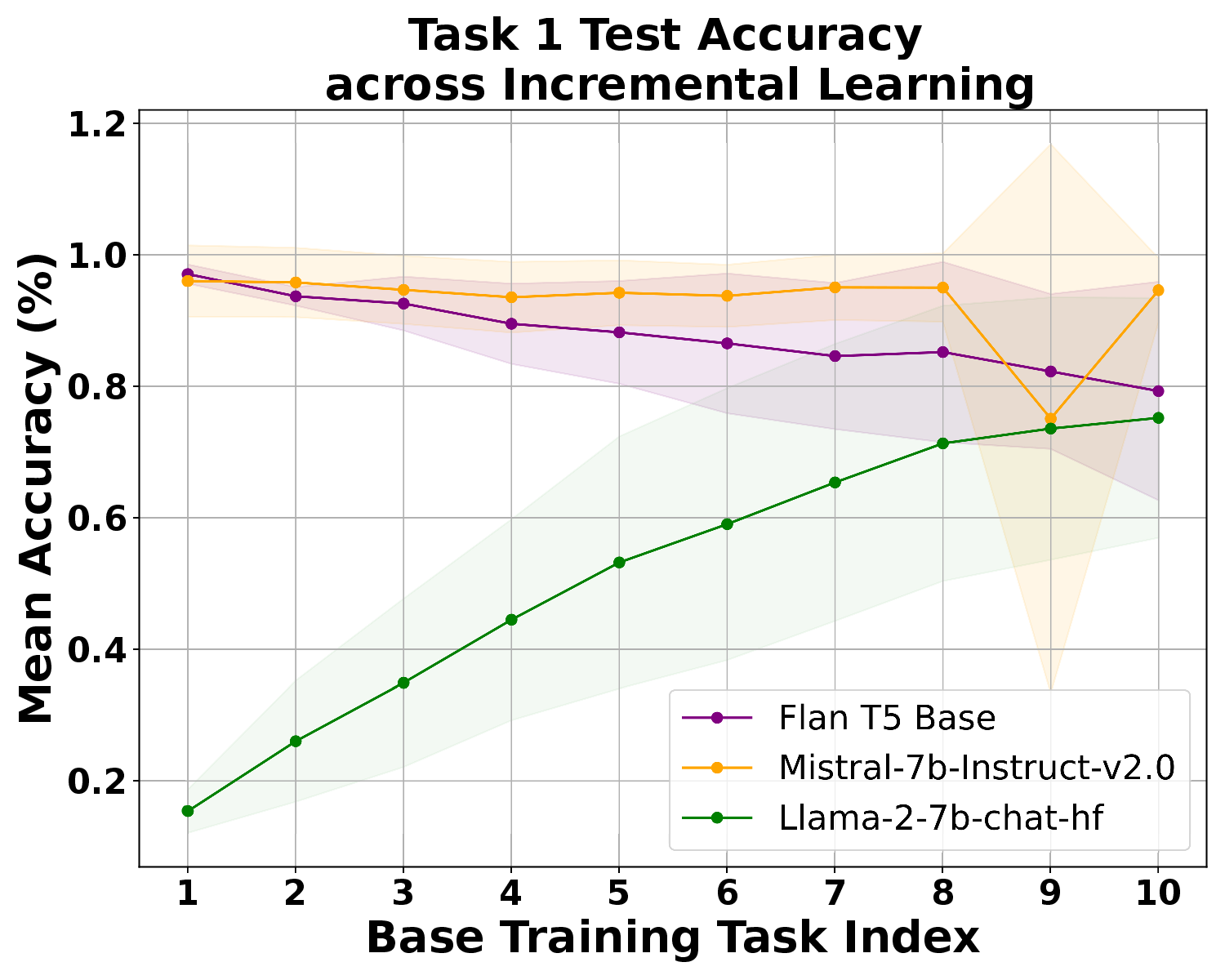}
        \caption{FewRel}
        \label{fig:fewrel}
    \end{subfigure}
    \caption{Mean test accuracies for Task 1 across five incremental learning runs with three language models are shown, with shaded areas representing the standard deviation.}
    \label{fig:comparison}
\end{figure}

\begin{figure}[H]
\centering
    % T5 Subfigure
    \begin{subfigure}[b]{0.4\textwidth}
        \centering
        \includegraphics[width=\linewidth]{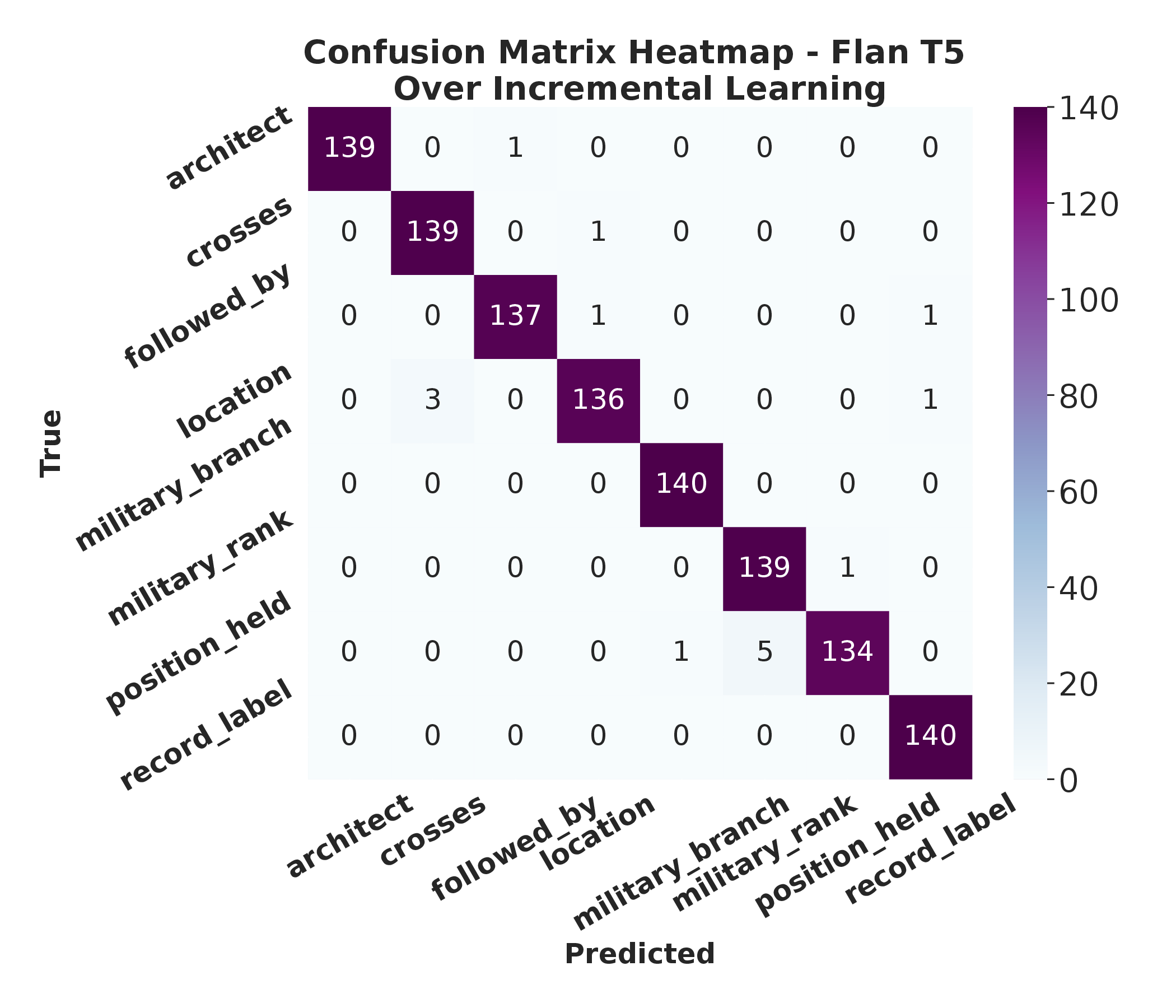}
        \caption{Base Training on Task 1}
        \label{fig:task_1_t5}
    \end{subfigure}
    \begin{subfigure}[b]{0.4\textwidth}
        \centering
        \includegraphics[width=\linewidth]{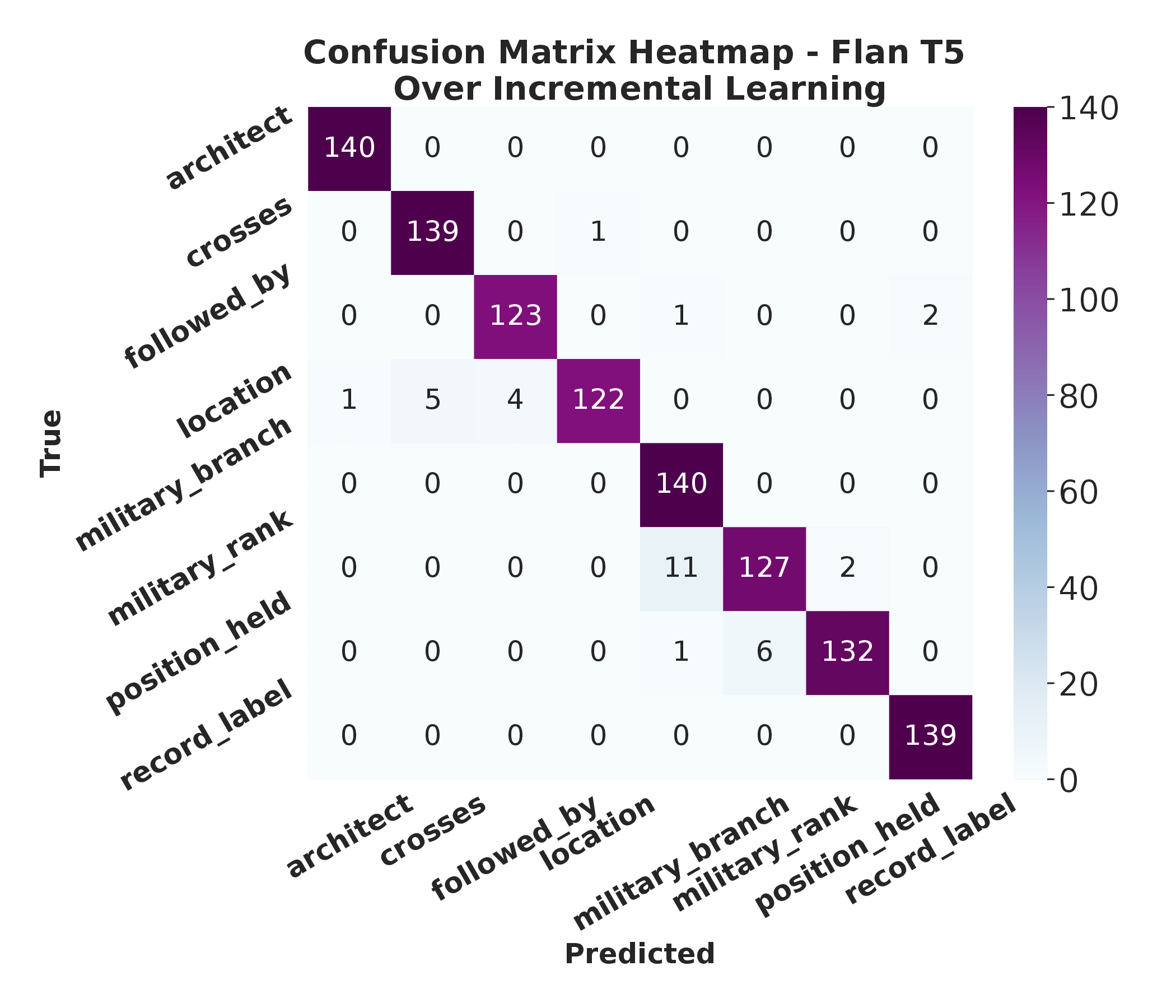}
        \caption{Base Training on Task 10}
        \label{fig:task_10_t5}
    \end{subfigure}
    
    \caption{Confusion Matrices for Task 1 and Task 10 in FewRel during Incremental Learning (run 1) with Flan-T5 Base.}
    \label{fig:confusion_matrix_t5}
\end{figure}

\begin{figure}[H]
    % \ContinuedFloat
    \centering
    % Llama2 Subfigure
    \begin{subfigure}[b]{0.4\textwidth}
        \centering
        \includegraphics[width=\linewidth]{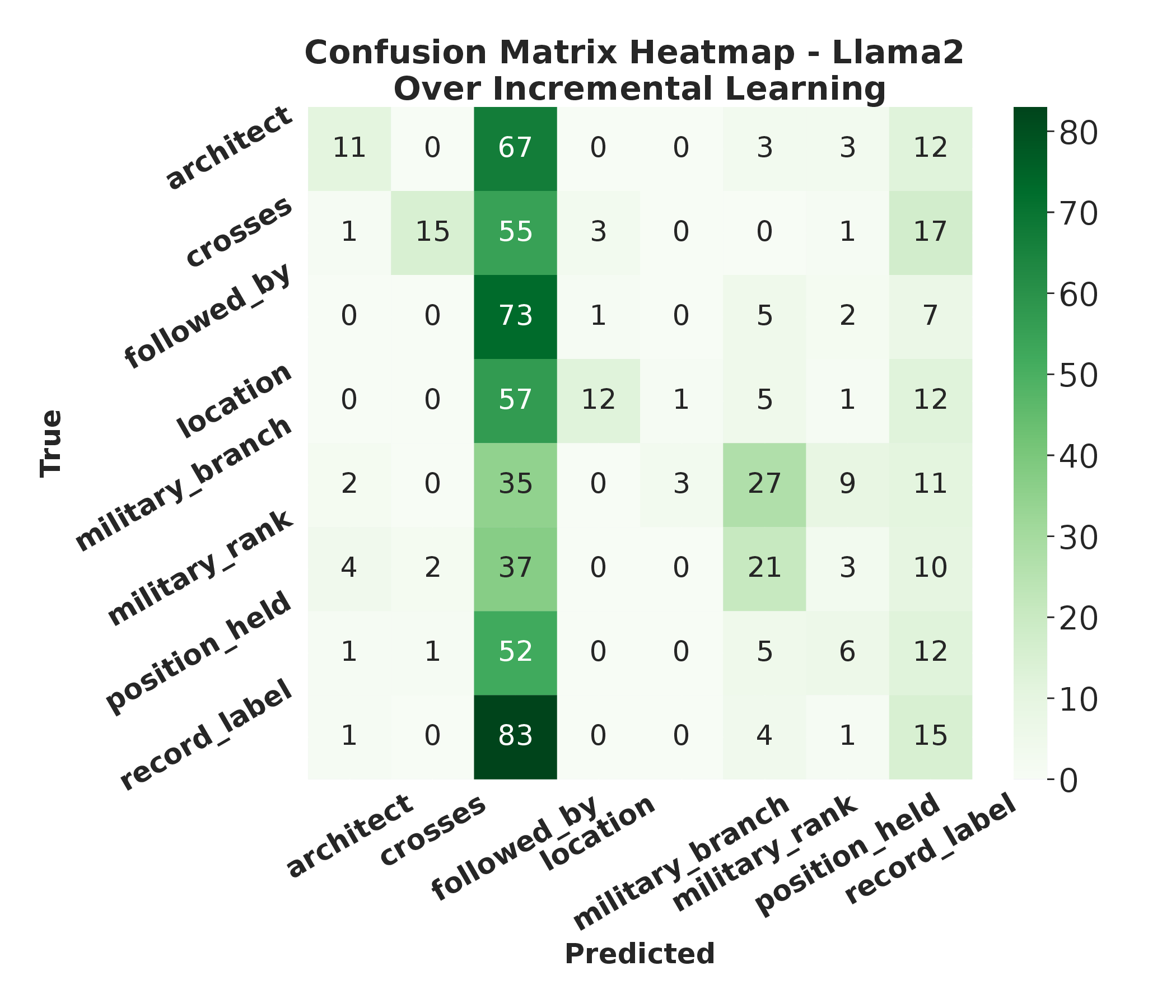}
        \caption{Base Training on Task 1 (Llama2)}
        \label{fig:task_1_llama}
    \end{subfigure}
    \begin{subfigure}[b]{0.4\textwidth}
        \centering
        \includegraphics[width=\linewidth]{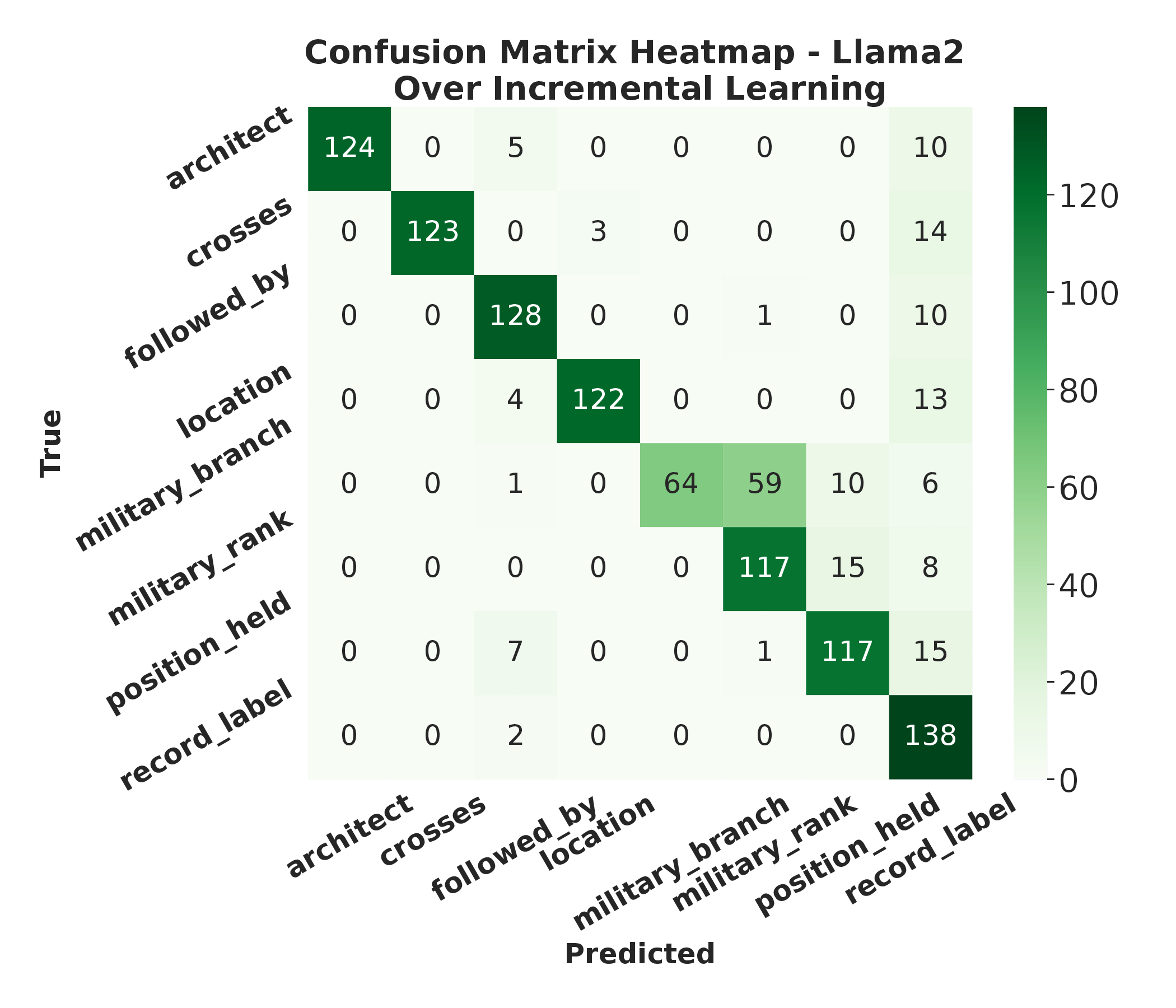}
        \caption{Base Training on Task 10 (Llama2)}
        \label{fig:task_10_llama}
    \end{subfigure}
    
    % Mistral Subfigure
    \begin{subfigure}[b]{0.4\textwidth}
        \centering
        \includegraphics[width=\linewidth]{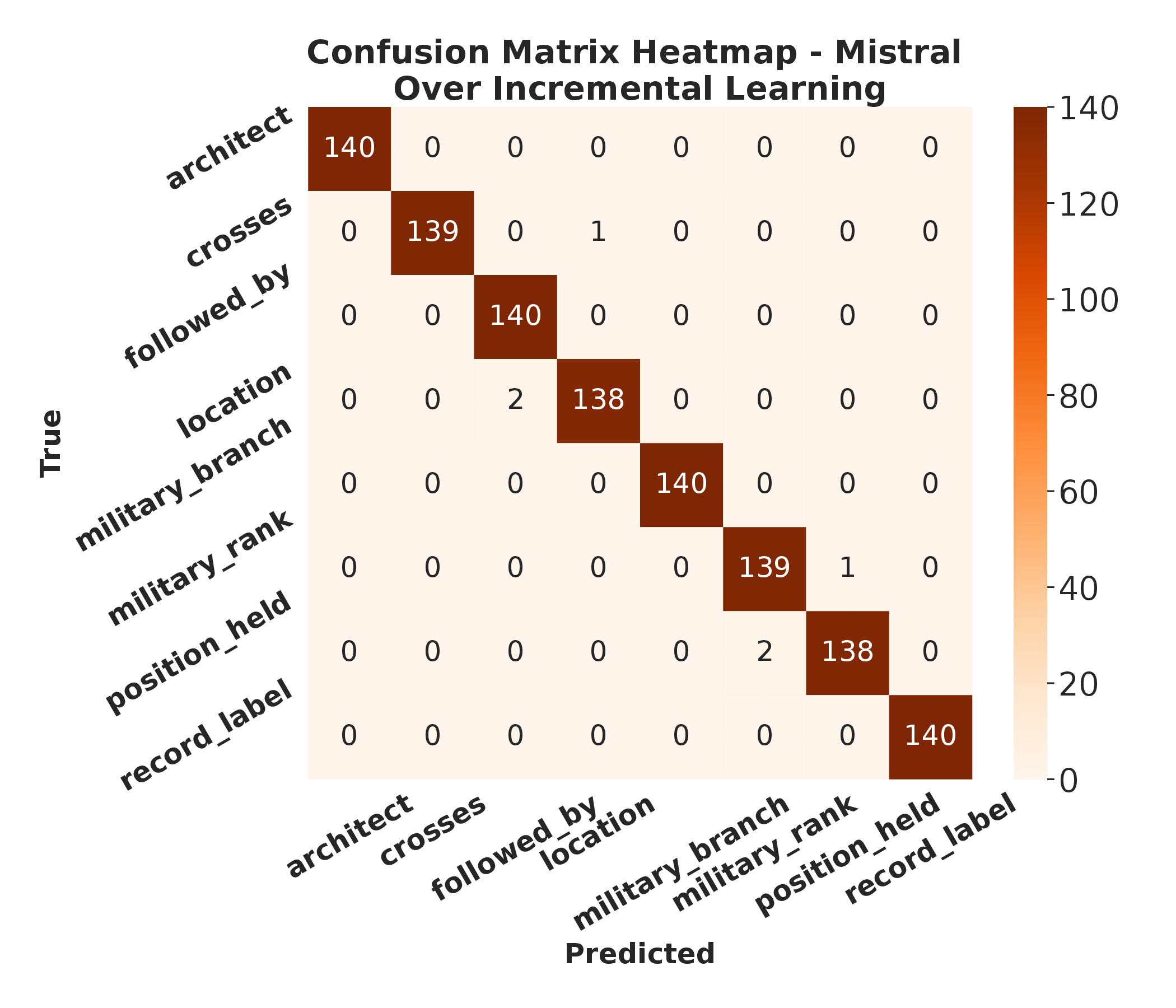}
        \caption{Base Training on Task 1 (Mistral)}
        \label{fig:task_1_mistral}
    \end{subfigure}
    \begin{subfigure}[b]{0.4\textwidth}
        \centering
        \includegraphics[width=\linewidth]{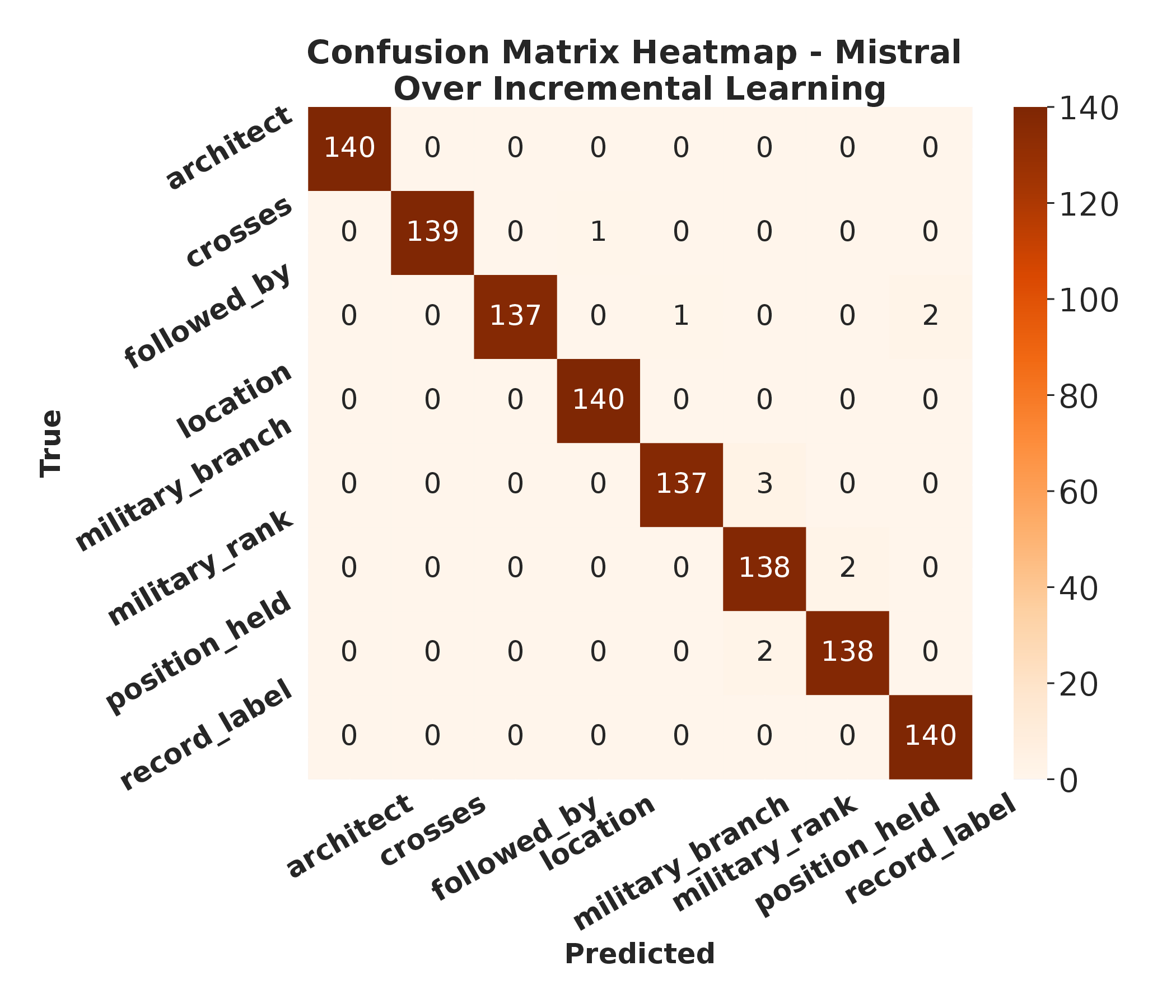}
        \caption{Base Training on Task 10 (Mistral)}
        \label{fig:task_10_mistral}
    \end{subfigure}
    \caption{Confusion Matrices for Task 1 and Task 10 in FewRel during Incremental Learning (run 1) with Llama2 and Mistral.}
    \label{fig:mist_llama}
\end{figure}

In this section, we analyze and visualize knowledge transfer in incremental task learning (ITL). To assess knowledge transfer across ITL, we examine the mean test accuracy of Task 1 from both FewRel and TACRED, utilizing three language models, as illustrated in~\Cref{fig:comparison}. The  mean test accuracy of Task 1 shows a slight decline as Flan T5 Base progresses from training on Task 1 to Task 10, where TACRED is evaluated. Similarly, Mistral also experiences a bit forgetting on this dataset. In contrast, Llama2 achieves positive backward knowledge transfer over ITL, which enhances the model's performance on earlier tasks (e.g., Task 1, as indicated in~\Cref{fig:tacred}). Furthermore, we analyze the performance of these language models on the FewRel dataset as well. Flan T5 Base suffers from significant catastrophic forgetting throughout ITL, as depicted in~\Cref{fig:fewrel}. We illustrate Flan T5 Base's behavior with a confusion matrix in~\Cref{fig:confusion_matrix_t5}, which highlights how test dataset of Task 1 performs during ITL. Flan T5 Base performs better on the Task 1's test dataset after training on the Task 1's training set (see~\Cref{fig:task_1_t5}); however, it generates hallucinated predictions—responses not among the predefined relation types—when encountering catastrophic forgetting, as shown in~\Cref{fig:task_10_t5}. Note that the FewRel test dataset contains 140 samples per relation type. While calculating the number of predictions in the confusion matrices (see~\Cref{fig:mist_llama,fig:confusion_matrix_t5}), we exclude statistics on hallucinated predictions. Similar to the TACRED results, Mistral faces forgetting issues on FewRel, as illustrated in~\Cref{fig:fewrel}. Llama2 demonstrates comparable positive backward knowledge transfer on FewRel (see~\Cref{fig:fewrel}), while reducing the hallucinated predictions by the completion of ITL. This reduction is evident from the decrease in false predictions throughout ITL, as shown in~\Cref{fig:task_1_llama,fig:task_10_llama}. In conclusion, language models may generate the hallucinated predictions when they fail to transfer previously learned knowledge forward. Additionally, the number of hallucinated predictions tends to decrease when backward knowledge transfer occurs, as observed with Llama2.


\section{Ablation Study}
\label{sec:ablation_study}
\begin{table}[H]
\tiny
\centering
\caption{Mean Average Accuracy (\textit{a}), Whole Accuracy (\textit{w}) (\%) and Backward Knowledge Transfer (bwt) (\%) on the TACRED dataset with memory sizes (\textit{m}).}

\sisetup{table-format=3.1} % Configures decimal alignment
\begin{tabular}{ccc|ccc|ccc|ccc}
\toprule
\multicolumn{3}{p{3cm}}{\textbf{without memory replay (m = 0)}} & \multicolumn{3}{c}{\textbf{\textit{m} = 5}} & \multicolumn{3}{c}{\textbf{\textit{m} = 10}} & \multicolumn{3}{c}{\textbf{\textit{m} = 15}} \\
\cmidrule(lr){1-3} \cmidrule(lr){4-6} \cmidrule(lr){7-9} \cmidrule(lr){10-12}
\textbf{\textit{w}} & \textbf{\textit{a}} & \textbf{\textit{bwt}} & \textbf{\textit{w}} & \textbf{\textit{a}} & \textbf{\textit{bwt}} & \textbf{\textit{w}} & \textbf{\textit{a}} & \textbf{\textit{bwt}} & \textbf{\textit{w}} & \textbf{\textit{a}} & \textbf{\textit{bwt}} \\
\midrule
92.7 & 92.1 & -1.319 & 95.2 & 95.1 & -0.276 & 95.8 & 95.8 &- 0.204 & 96.3 & 96.3 & 0.283 \\
\bottomrule
\end{tabular}
\label{tab:memory_experiments}
\end{table}
We examine how varying memory sample sizes (e.g., 5, 10, 15) affect incremental task fine-tuning using the Flan-T5 Base model on the TACRED dataset (see~\Cref{tab:memory_experiments}).
% Larger memory sizes enhance performance on seen tasks, while mitigating catastrophic forgetting during incremental training across ten tasks. 
We perform a two-tailed significance test with \( H_0 \): no difference between the condition without memory replay and memory size m, and \( H_1 \): a significant difference between m=0 (no replay) and m= 5, 10 or 15. The results show significant differences in memory sizes, with p-values of 0.0958 (m= 0 vs 5), 0.0509 (m= 0 vs 10), and 0.0262 (m= 0 vs 15), all below $\alpha$=
0.10 (level of significance), leading to the rejection of \( H_0 \).

\section{Discussion}  
\label{sec:discussion}  
Continual relation extraction (CRE) has traditionally focused on incremental task learning, a subset of continual learning. However, existing methods often struggle with forward knowledge transfer, leading to catastrophic forgetting. Memory replay has proven to be an effective mitigation strategy, yet it offers limited adaptability during incremental learning and fails to sufficiently minimize forgetting~\cite{chen-etal-2023-consistent,cui-etal-2021-refining,wang-etal-2019-sentence,ye-etal-2024-distilling,zhang_2022}.

In this work, we investigate whether pre-trained language models influence knowledge transfer in CRE, seeking an answer to the research question: \textit{To what extent do pre-trained language models affect knowledge transfer, including backward transfer, in continual relation extraction?} Positive backward knowledge transfer is observed with Llama2 on both datasets, reducing false predictions and hallucinations on earlier tasks in~\Cref{sec:kta}. In contrast, Flan T5 Base generates hallucinated predictions, particularly when encountering significant knowledge forgetting (see~\Cref{fig:confusion_matrix_t5}). While custom CRE models produce false predictions among predefined relation types, language models tend to generate hallucinations beyond these predefined types,  even when fine-tuned for a specific task.

Additionally, we also compare our findings with the state-of-the-art (SoTA) method~\cite{zhang_2022}. Mistral outperforms Flan T5 Base and Llama2 on both datasets across evaluation metrics, although it exhibits slight forgetting on FewRel (see~\Cref{tab:w_a_acc}). Furthermore, it surpasses previously published SoTA results (\cite{zhang_2022}) on TACRED in terms of all evaluation metrics (see~\Cref{tab:results,tab:w_a_acc}) and exceeds KIP’s performance on FewRel in terms of seen task accuracy (see~\Cref{tab:results}) at the end of ITL. However, Mistral ranks second when considering whole and average accuracies in~\Cref{tab:w_a_acc}. Notably, KIP leverages prompt-based approaches and multi-head scaled dot-product attention alongside memory replay.

\section{Conclusion}
\label{sec:conclusion}
This work evaluates three large language models—Flan T5, Mistral and Llama2— for incremental task learning in continual relation extraction on the FewRel and TACRED datasets, with a focus on knowledge transfer and catastrophic forgetting. Despite memory replay, Flan T5 suffers from significant forgetting, though an ablation study confirms the effectiveness of memory replay and identifies optimal memory configurations. Mistral exhibits slight forgetting on FewRel but achieves positive backward knowledge transfer on TACRED, surpassing previous works and achieving state-of-the-art performance in seen task accuracy on these datasets. Llama2 demonstrates consistent positive knowledge transfer on both datasets. Pretrained models like Mistral outperform custom models. 
The weakness of this approach is the natural tendency for hallucinated predictions caused by catastrophic forgetting. In future work, to tackle this problem, we aim to apply fact checking and retrieval-augmented generation by incorporating information about entities from knowledge bases, e.g., Wikidata, into the prompt template during the test phase.
% \newpage
\bibliographystyle{splncs04}
\bibliography{references}
% \newpage

\end{document}